\documentclass[letterpaper, 10 pt, conference]{ieeeconf}
\usepackage{graphicx}
\usepackage{tcolorbox}
\AtBeginDocument{%
  \providecommand\BibTeX{{%
    \normalfont B\kern-0.5em{\scshape i\kern-0.25em b}\kern-0.8em\TeX}}}

\usepackage{graphicx}
\usepackage{wrapfig}
\usepackage{booktabs} 

\usepackage{url}
\usepackage{algorithm}
\usepackage[noend]{algpseudocode}
\usepackage{amssymb,amsmath}
\usepackage{subcaption}
\usepackage{graphicx}
\usepackage{comment}

\usepackage{amsthm}
\usepackage{thmtools, thm-restate}
\usepackage{multirow}

\usepackage[graphicx]{realboxes}

\usepackage{xcolor} 
\definecolor{ocre}{RGB}{243,102,25} 

\usepackage{avant} 
\usepackage{mathptmx} 

\usepackage{microtype} 
\usepackage[utf8x]{inputenc} 

\usepackage{tikz}
\usetikzlibrary{shapes}
\RequirePackage{ifthen}
\RequirePackage{amsmath}
\RequirePackage{amsthm} 
\RequirePackage{amssymb}
\RequirePackage{xspace}
\RequirePackage{graphics}
\usepackage{color}
\usepackage{keyval}
\usepackage{xspace}
\usepackage{mathrsfs}
\newcommand{\hussein}[1]{\textcolor{black}{#1}}
\newcommand{\yuxuan}[1]{\textcolor{black}{#1}}

\newcommand{\dom}{\relax\ifmmode {\mathit{dom}} \else ${\sf dom}$\fi}

\usepackage{setspace}
\usepackage{listings}

\newcounter{theorems}
\newtheorem{theorem}[theorems]{Theorem}

 \newcounter{definitions}
 \newtheorem{definition}[definitions]{Definition}

\usepackage[normalem]{ulem}

\usepackage{placeins}
\makeatletter
\newcommand\nocaption{%
    \renewcommand\p@subfigure{}
    \renewcommand\thesubfigure{\thefigure\alph{subfigure}}
}
\makeatother
\usepackage{makecell}
\usepackage{svg}
\title{
Pre-Trained Vision Models as Perception Backbones 
for Safety Filters in Autonomous Driving}
\author{
	Yuxuan Yang and Hussein Sibai \\
	Computer Science and Engineering Department\\
	Washington University in St. Louis \\
	\texttt{\{y.yuxuan,sibai\}@wustl.edu} \\
}

\begin{document}

\maketitle


\begin{abstract}
\hussein{
End-to-end vision-based autonomous driving has achieved impressive success, but safety remains a major concern. The safe control problem has been addressed in low-dimensional settings using safety filters, e.g., those based on control barrier functions. Designing safety filters for vision-based controllers in the high-dimensional
settings of autonomous driving can similarly alleviate the safety problem, but is significantly more challenging. In this paper, we address this challenge by using frozen pre-trained vision representation models as perception backbones to design vision-based safety filters, inspired by these models' success 
 as backbones of robotic control policies. 
We empirically evaluate the offline performance of four common pre-trained vision models in this context. We try three existing methods for training safety filters for black-box dynamics, as the dynamics over representation spaces are not known.
We use the DeepAccident dataset that consists of action-annotated videos from multiple cameras on vehicles in CARLA simulating real accident scenarios. Our results show that the filters resulting from our approach are competitive with the ones that are given the ground truth state of the ego vehicle and its environment.} 
\end{abstract}


\section{Introduction}

\hussein{Computer vision plays a critical role in autonomous driving (AD)
~\cite{tesla2024blog,waymo2020blog}.
Videos provide detailed information about the vehicle and its environment at a low cost, leading, for example, Tesla to adopt a purely vision-based autonomous driving~\cite{tesla2024blog}. However, safety of vision-based driving policies remains the main concern that limits their wider deployment. Several approaches have been proposed to address this challenge besides training on larger datasets, including formal verification~\cite{Verification_of_image_based_NN_using_generative_models_JAIS_2022,Vision_Based_Autonomous_Aircraft_Landing_Yasser_NFM_2022,vision_controllers_verification_Chiao_Sayan_2022,Closed_Loop_Analysis_of_Vision_Based_Autonomous_Systems_CAV_2023,vision_based_verification_chuchu_stanley_2024}, online monitoring~\cite{monitor_ensembles_Hazem_Torfah_RV_2023,monitoring_using_LLMs_Marco_Pavone_2023},  and safe reinforcement learning~\cite{Safe_RL_survey_2015,safe_RL_for_AV_CPO_ITSC_2020,coptidice_lee2022,CPO_abbeal_ICML_2017,PPO_Safe_2017}.}

\hussein{Separately, control barrier functions (CBF) have been used to formally guarantee safety of control systems~\cite{CBF_survey_2019,CBF_and_Input_to_State_Safety_for_AD_TCST_2023}.
They can be used to construct constraints that distinguish safe controls from unsafe ones at any given state. One can then formulate a constrained optimization problem to generate controls that minimally deviate from nominal ones while maintaining safety~\cite{cbf}. Thus, a CBF can play the role of a {\em safety filter} that adjust a nominal control input as needed to maintain the safety of the system. 
When the dynamics of a control system are known,  its state is low-dimensional, and the set of safe states is formally defined, we can, under additional conditions, synthesize corresponding CBFs using methods like Sum-of-Squares (SoS)~\cite{Verification_and_Synthesis_using_SoS_Andrew_Clark_CDC_2021,Permissive_CBF_SOS_2018_Magnus} and Hamilton-Jacobi reachability analysis~\cite{refining_CBF_using_hamilton_jacobi_IROS_2022,robust_CBF_for_safety_critical_control_Sylvia_Herbert_CDC_2021}. However, in the complex and dynamic environments of AD, the combined state of the vehicle and its environment is high-dimensional. Moreover, formally defining the sets of safe and unsafe states and the  dynamics over such a high-dimensional space is infeasible. This limits the ability of traditional control methods to 
design of corresponding controllers and safety filters.
Recent deep learning-based approaches have  
achieved impressive success in the past few years addressing such limitations~\cite{end_to_end_AD_survey_2024,Vision_based_AD_using_deep_learning_survey_2022}. }

\hussein{Deep learning approaches for training neural driving policies are either end-to-end or modular. The former ones 
train models that directly map sensor observations to control inputs, limiting the information loss 
that usually results from 
modular approaches~\cite{end_to_end_AD_survey_2024}.
For vision-based safety filters, most existing methods follow a modular approach. 
They consider perception models that map images to interpretable low-dimensional states and systems with white-box, i.e., known, dynamics over these states~\cite{NeRF_CBF_tong_chuchu_icra_2023,vision_based_CBF_sarah_dean_CoRL_2021,vcbf,BarrierNet_DanielaRus_2023}. Few works considered the case of black-box dynamics, e.g.,~\cite{sablas,hyperplane,indcbf}, which can be suitable for non-interpretable state spaces' settings
and potentially be used to train end-to-end filters. 
} 

\hussein{Recently, pre-trained vision representation models (PVRs)\footnote{We use the same acronym as \cite{unsurprising}.}, which are trained in a self-supervised manner on large datasets \cite{CLIP,mae}, have been shown to improve the performance and sample complexity of end-to-end control policies in various control  tasks~\cite{unsurprising,vc1_2023,R3M_CoRL_2022,MVP_2022}. Such models learn to process images into low-dimensional, non-interpretable, vectors while preserving essential semantic features. A policy can then be trained on task-specific data to map these features to low-level control inputs, without having to learn common image processing skills from the scarce and costly robotic task-specific data, which it would have to do otherwise. 
}
\hussein{In this paper, we empirically evaluate the effectiveness of four common PVRs when used as backbones of vision-based safety filters, instead of policies, for AD. We define safety as collision avoidance and use the recently released DeepAccident~\cite{deepaccident} dataset to train our safety filters. It consists of collision and collision-free trajectories of vehicles simulated using CARLA~\cite{Carla}. The trajectories consist of multi-camera action-annotated videos.  }
We 
\hussein{experiment with} three different \hussein{training} methods  \hussein{for safety filters for black-box control systems.} 
\hussein{We also evaluate different} methods \hussein{to train safety filters on multi-view camera feeds.}
Our 
results can be summarized as follows: \yuxuan{(1) PVRs pre-trained with 
control-related datasets can help 
achieve better safety filtering performance \hussein{than others, }
(2) 
\hussein{filters} trained 
\hussein{on top of PVRs} 
are able to approach or even sometimes outperform filters with access to   
ground truth states, (3) 
a recent method that trains filters for black-box systems that directly map states to classifiers in the control space outperforms  those that learn dynamics first in terms of detecting unsafe control inputs, 
and (4) in a multi-camera feed setting, fusing the representations into a single  state 
is better than having a separate 
state for each camera feed and learning separate dynamics.  
}

\section{Related Work}
\label{sec:RelatedWork}
\noindent \textbf{Vision-based safety filters.} BarrierNet~\cite{BarrierNet_DanielaRus_2023} introduces a differentiable layer that can be added on top of a vision-based neural controller to guarantee safety, but requires known dynamics and safety constraints. NeRF-based CBFs~\cite{NeRF_CBF_tong_chuchu_icra_2023} define CBFs based on the depth dimension of RGBd images and use  
neural radiance field models to map predicted robot's positions, using known dynamics, to images. The authors in \cite{vision_based_CBF_sarah_dean_CoRL_2021} propose an approach to design CBFs for known systems that are robust to errors generated by a perception model that maps images to states.  V-CBF~\cite{vcbf}, defines vision-based CBFs by assuming the existence of a model that segments images into safe and unsafe regions and a C-GAN 
that computes the gradient of the segmented images with respect to the robot's position. Its dot product with 
the known dynamics produces the needed time derivative of the CBF. 


\noindent \textbf{PVRs as backbones for control.} Several works showed the benefits of using PVRs as backbones for control policies. The authors in  \cite{unsurprising} showed that frozen PVRs as backbones of control policies is competitive with, and can sometimes outperform, policies with access to ground truth states in various benchmarks.   
The authors in \cite{vc1_2023} also evaluated different PVRs as backbones, both frozen and finetuned, for policies to check if there is one that is dominant on 17 different control tasks, finding none, and present a new one, VC1, that outperforms them on average, but not dominant either.
These works focus on training control policies instead of safety filters.
	
\section{Preliminaries}
\label{sec:Preliminaries}
\subsection{Control barrier functions}
Control barrier functions (CBFs) are usually designed to act as safety filters for continuous-time nonlinear control-affine systems. A  nonlinear control-affine system has the form of:
\begin{align}
\label{eq:system}
    \dot{x}=f(x)+g(x)u,
\end{align}
where $x\in X \subset \Bbb{R}^n$ is the system state and $u\in U\subset \Bbb{R}^m$ is the control input. We also assume that $f:X\rightarrow \Bbb{R}^n$ and $g:X\rightarrow \Bbb{R}^{n\times m}$ are locally Lipschitz continuous functions and that system~(\ref{eq:system}) is forward complete. 

\begin{definition}[Zeroing control barrier functions \cite{cbf}]
    Consider system (\ref{eq:system}) and a set of states $X_{\mathit{safe}}\subset X$ that we consider safe. A continuously differentiable function $B:X\rightarrow \Bbb{R}$ is a {\em zeroing control barrier function} \hussein{for system~(\ref{eq:system})} if it satisfies the following conditions:
    \begin{equation}\label{eq:cbf}
        \begin{aligned}
            B(x)&\geq 0 \quad \forall x\in X_{\mathit{safe}},\\
            B(x)&< 0 \quad \forall x\in X\setminus X_{\mathit{safe}}, \\
            \exists u\in U \text{ s.t. } \dot{B}(x,u)+\gamma(B(x))&\geq 0 \quad \forall x\in X,\\
        \end{aligned}
    \end{equation}
where \hussein{$\dot{B}(x,u)$ is the time derivative of $B$ along the trajectories of (\ref{eq:system}), which is equal to $\nabla B(x) (f(x) + g(x) u)$ by the chain rule, and} $\gamma:\Bbb{R}\rightarrow \Bbb{R}$ is an extended class $\kappa_\infty$ function, i.e.,  
strictly increasing, $\gamma(0)=0$, and $\lim_{x\rightarrow \infty} \gamma(x)=\infty$.
\end{definition}

\begin{theorem}
Any Lipschitz continuous control policy 
\hussein{$\pi: X \rightarrow U$} that satisfies 
\begin{align}
\label{eq:CBF_set}
    \pi(x)\in\{u\in U:\nabla B(x)(f(x)+g(x)u)+\gamma(B(x))\geq 0\}
\end{align}
will  
\hussein{make $X_{\mathit{safe}}$}
a forward invariant set \cite{cbf}. 
\end{theorem}

Then, given 
\hussein{a} reference controller \hussein{ $\pi_{\mathit{ref}}:X\rightarrow U$, }
the CBF $B$ can serve as a safety filter \hussein{that corrects $\pi_{\mathit{ref}}$'s decisions when violating safety constraints. The safety filter can be constructed by formulating a quadratic program (QP) with the objective of finding a control input that is minimally distant from the reference one while belonging to the set  in (\ref{eq:CBF_set}) that guarantees safety. The QP problem is as follows~\cite{cbf}: }
\begin{equation}\label{eq:qp}
    \begin{aligned}
    \pi_{\mathit{safe}}(x) &:= \arg\min_{u\in U}\lVert u-\pi_{\mathit{ref}}(x) \rVert^2\\
        s.t.&\quad \nabla B(x)(f(x)+g(x)u)+\gamma(B(x))\geq 0. 
    \end{aligned}
\end{equation}

\subsection{Safety filters for black-box systems}
\hussein{In our setting, we consider the state of the system composed of a vehicle and its environment to be a learned function of the representations produced by the PVRs of the frames that is not interpretable. 
The dynamics of such a system over such a state space are unknown.
We thus have to rely on the methods designed to train safety filters for black-box dynamics to train ours.}
Here, we \hussein{describe} 
three \hussein{existing methods for such a setting.} 
Two of them are CBF-based, which require \hussein{us to train} a dynamical model to \hussein{approximate the ground-truth one.} 
The \hussein{third 
one trains a model that maps states to}  
hyperplanes \hussein{that separate safe and unsafe control inputs in an end-to-end manner, without training an approximate dynamical model.}

\subsubsection{In-Distribution Barrier Function (iDBF) \hussein{\cite{indcbf}}}
iDBFs \hussein{were introduced
as an application of CBF theory to the problem of out-of-distribution states avoidance when learning controllers  from  an offline dataset of trajectories. It} 
\hussein{assumes the trajectories in the training dataset are expert demonstrations and thus considers all the states they visited as}
safe. 
It 
\hussein{trains a control-affine dynamical model}
\hussein{paramaterized by $\theta$:} 
$\dot{x}=f_{\theta}(x)+g_{\theta}(x)u$, and 
\hussein{a CBF} $B$ \hussein{parameterized by $\phi$:} 
$B_{\phi}:X\rightarrow \Bbb{R}$. 
\hussein{Both are trained using}  
a self-supervised learning approach. \hussein{While the dynamics are continuous-time ones, the images are captured at sampled time instants. Thus, the iDBF method trains $f_\theta$ and $g_\theta$ as  neural ODEs  \cite{neuralODE}.} 
For a given time step $t$ at which an image $I_t$ is captured,  
the  
state $x_t$ is encoded via \hussein{an auto-encoder model consisting of the pair $E_\psi$ and $D_\psi$ of encoder and decoder networks, respectively. The encoder $E_\psi$}
takes 
$I_t$, the previous state $x_{t-1}$, and the previous control $u_{t-1}$ 
as input: $x_t = E_\psi(I_t,x_{t-1},u_{t-1})$.
\hussein{For $B_{\phi}$ to be an iDBF, it has to assign non-negative values to safe states, i.e., those in the training dataset distribution, and}
\hussein{negative values for other states. It should also satisfy the third condition in (\ref{eq:cbf}) for state-control pairs that appear in the training dataset as they map in-distribution states to in-distribution ones.} 
\hussein{Accordingly, $B_{\phi}$ is trained using the following loss function: }
\begin{align}
   \label{eq:idbf_loss_function}
    L_{iDBF} &:= \frac{w_{\mathit{safe}}}{N_{\mathit{safe}}}\sum_{x_{\mathit{safe}}}\sigma(\epsilon_{\mathit{safe}}-B_{\phi}(x_{\mathit{safe}})) \nonumber\\
        +& \frac{w_{\mathit{unsafe}}}{N_{\mathit{unsafe}}}\sum_{x_{\mathit{unsafe}}}\sigma(\epsilon_{\mathit{unsafe}}+B_{\phi}(x_{\mathit{unsafe}}))\nonumber\\
        +& \frac{w_{ascent}}{N_{\mathit{safe}}}\sum_{x_{\mathit{safe}}}\sigma(\epsilon_{ascent}-\nabla B_{\phi}(x_{\mathit{safe}})(f_\theta(x_{\mathit{safe}}) \nonumber\\
        &+g_\theta(x_{\mathit{safe}})u_{\mathit{safe}})-\gamma(B_\phi(x_{\mathit{safe}}))),
\end{align}
where $\sigma(\cdot)$ is the ReLU activation function and  $\{(x_{\mathit{safe}},u_{\mathit{safe}})\}$ are sampled from the state-control pairs in the 
trajectories in the training dataset. The constants  $\epsilon_{\mathit{safe}}$,$\epsilon_{\mathit{unsafe}}$, and $\epsilon_{ascent}$ are hyper-parameters that \hussein{guide the training of the barrier function to assign values that robustly satisfy the} 
conditions in (\ref{eq:cbf}). To obtain samples of out-of-distribution 
states
$x_{\mathit{unsafe}}\in X\setminus X_{\mathit{safe}}$, the authors trained a multi-modal Gaussian distribution over \hussein{controls} 
conditioned on the state with density $\pi_{BC}(u|x)$. Then, during the training, for each $x_{\mathit{safe}}$ they sample $N_{candidate}$ control inputs 
and select $u_{candidate}$, one \hussein{that $\pi_{BC}$ assigns a} 
value below a pre-defined threshold. Then, they simulate the learned dynamical model for one time step following the control input 
\hussein{and consider the last state as out-of-distribution, or equivalently in their setting, unsafe.}
\hussein{In our case, we are not addressing the out-of-distribution states avoidance problem. Instead, we have safe and unsafe (sub)trajectories in the dataset, and thus do not need the second part of the method to generate unsafe ones. We thus just replace the safe and unsafe states and controls in (\ref{eq:idbf_loss_function}) with those with the same label in our considered dataset.}  
Given a uniformly time-sampled sequence of images of length $T$, the training loss function for the dynamics  \hussein{and the auto-encoder} is defined as follows:
\begin{align}\label{eq:lossdyn}
    L_{dyn}:=\frac{1}{T}\sum_{t=1}^T w_{1}\lVert \tilde{x}_{t}-x_t\rVert^2+w_{2}\lVert \tilde{I}_{t}-I_t\rVert^2+w_{3}\lVert \hat{I}_{t}-I_t\rVert^2,
\end{align}
where \yuxuan{$x_t$ is the output of $E_{\psi}$,} 
 $\Tilde{x}_t$ is obtained by \hussein{simulating the dynamics defined by $f_\theta$ and $g_\theta$ starting from $x_{t-1}$ and following $u_{t-1}$ as a constant control for the duration of the sampling time,} 
$\Tilde{I}_t$ is \hussein{an image generated by}
decoding 
$\Tilde{x}_t$ \hussein{using $D_\psi$}, and $\hat{I}_t$ is the encoded \hussein{the} decoded version of $I_t$.

\hussein{Given the trained iDBF, dynamical model, and auto-encoder, a quadratic program similar to the one in (\ref{eq:qp}) can be formulated and solved at sampling times. It is worth noting that solving (\ref{eq:qp}) and using the resulting control as a constant signal until the next sampling time does not necessarily guarantee safety as the behavior between the time steps has to be taken into consideration. Also, the learned dynamics and barrier are not guaranteed to accurately model the ground truth dynamics or satisfy the barrier conditions in (\ref{eq:cbf}), respectively. This is the case for all methods in this paper.} 

\yuxuan{Finally, since most of the existing PVRs have encoder-only architectures, we cannot use them to compute the terms the loss function in (\ref{eq:lossdyn}) that are based on decoding images from representations. Hence, in our implementation, we remove these terms,  resulting in the following simplified loss function: $L_{dyn}=\frac{1}{T}\sum_{t=1}^T \lVert \Tilde{x}_{t}-x_t\rVert^2.$}

\subsubsection{SABLAS~\cite{sablas}}
SABLAS is \hussein{a method to train safer controllers for}  
 black-box systems \hussein{by training them in parallel with CBFs}. It trains a dynamical model to approximate the 
 ground truth dynamics, which is similar to the iDBF approach. 
 \hussein{It assumes the availability of a simulator and easy-to-generate labels for  
 safe and unsafe states. } 
The main difference with the iDBF approach, besides the application to out-of-distribution avoidance versus safety, is that SABLAS 
\hussein{
accounts for the gap between the learned dynamical model and the ground truth during the training of the controller and the CBF, while iDBF does not. Also, SABLAS trains discrete-time dynamical models, instead of neural ODEs,  of the form:} 
\begin{equation}\label{eq:sablas}
    \begin{aligned}
    \Tilde{x}_{t+\Delta t} &= x_t + f_\theta(x_t,u_t)\Delta t,\\
    \Bar{x}_{t+\Delta t} &= \Tilde{x}_{t+\Delta t} + g(x_{t+\Delta t}-\Tilde{x}_{t+\Delta t}),
    \end{aligned}
\end{equation}
where $f_\theta:(X,U)\rightarrow X$ is 
parameterized with $\theta$, $x_{t+\Delta t}$ is obtained by 
using the black-box simulator to generate the actual trajectory 
\hussein{starting from} $x_t$ \hussein{and following the constant signal equal to $u_t$ for $\Delta t$ seconds}, and  $g(\cdot)$ is the detach function, i.e., an identity function that we treat as a constant instead of a function with a gradient, which is 
\hussein{called as} $g(x)=x.\text{detach}()$ in PyTorch \cite{pytorch}. This process enables \hussein{a better approximation of the gradient of the loss function used to train the CBF} 
when the black-box dynamics are not differentiable. 
SABLAS \hussein{defines the loss function for training $B_\phi$ as $L_{\mathit{SABLAS}} := \frac{w_{\mathit{safe}}}{N_{\mathit{safe}}}\sum_{x_{\mathit{safe}}}\sigma(\epsilon_{\mathit{safe}}-B_{\phi}(x_{\mathit{safe}}))+\frac{w_{\mathit{unsafe}}}{N_{\mathit{unsafe}}}\sum_{x_{\mathit{unsafe}}}\sigma(\epsilon_{\mathit{unsafe}}+B_{\phi}(x_{\mathit{unsafe}}))+\frac{w_{\mathit{ascent}}}{N_{\mathit{safe}}}\sum_{x_{\mathit{safe}}}\sigma(\epsilon_{ascent}-(B_{\phi}(\Tilde{x}_{\mathit{safe}})-B_{\phi}(x_{\mathit{safe}}))/\Delta t -\gamma(B_\phi(x_{\mathit{safe}}))),$}
where $\Delta t$ is the sampling period and $\Tilde{x}_{\mathit{safe}}$ is the predicted state starting from $x_{\mathit{safe}}$ after $\Delta t$ using  (\ref{eq:sablas}). \hussein{Once it learns $B_\phi$ and a controller (not discussed here) from a batch of trajectories, it generates another batch using the simulator following the trained controller. It then uses the trajectories to train the CBF and controller again using the same loss function, and repeat the process until convergence.} 
\yuxuan{
However, the code that was used to generate the DeepAccident dataset is not available publicly up to our knowledge. Thus, not being able to easily generate more trajectories, we only consider the first iteration of SABLAS that learns the CBF from the first set of trajectories, which for us are those in the DeepAccident dataset.} 

\begin{figure*}[hbt!]
    \centering
    \includegraphics[width=0.65\linewidth]{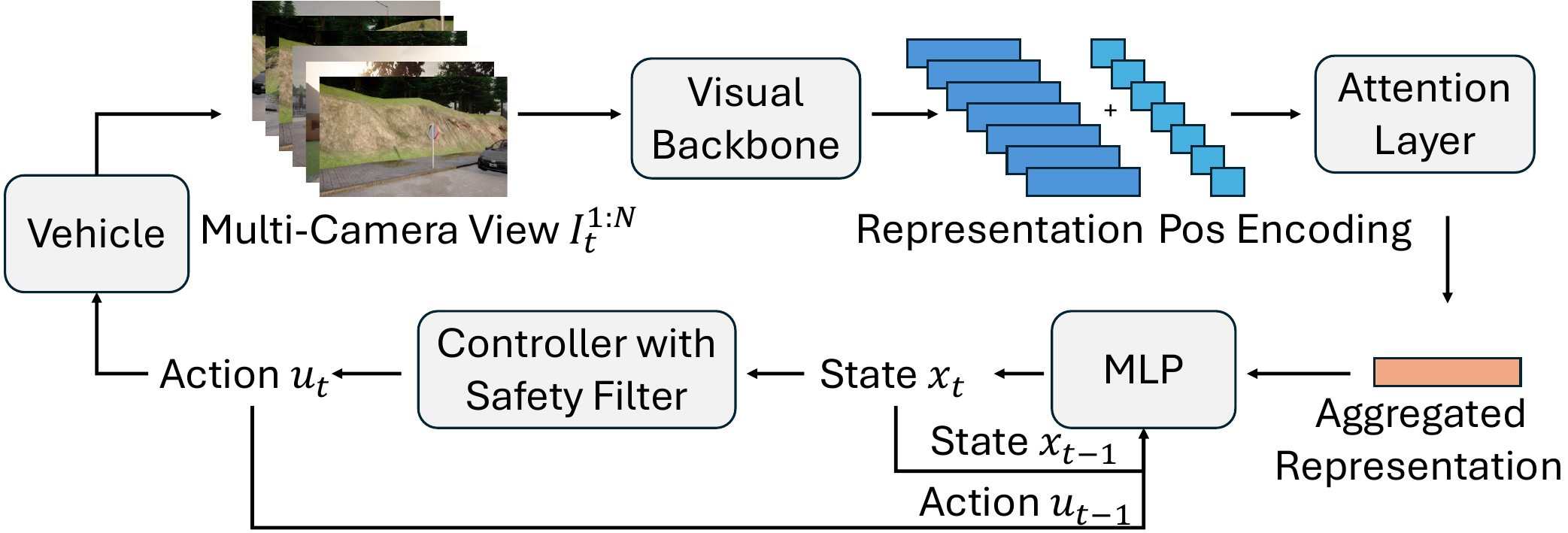}
    \caption{The inference diagram of vision-based safety filters designed using our approach.}
    \label{fig:framework}
\end{figure*}

\subsubsection{Safety Filters for Black-Box Dynamical Systems by Learning Discriminating Hyperplanes (DH) \cite{hyperplane}}
\hussein{The authors of \cite{hyperplane} proposed training a neural network that given a state, outputs a hyperplane that separates the safe control inputs from the unsafe ones  at that state. They generalized the third constraint of a CBF in (\ref{eq:cbf}), which also defines such a hyperplane when the dynamics are known, to the case when they are not. In contrast to the iDBF and SABLAS methods, this approach  does not train a dynamical model and directly trains the safety filter.} 
A \hussein{state-dependent} hyperplane 
over the control space $U$ at a state $x$ can be represented by the equation $a_\theta(x)^\top u=b_\theta(x)$, where $a_\theta: X \rightarrow U$ and $b_\theta: X \rightarrow \mathbb{R}$.
\hussein{
It can be used to adjust the} 
quadratic program in (\ref{eq:qp}) \yuxuan{by replacing the CBF-based constraint with $a_\theta(x)^\top u\geq b_\theta(x)$.}
It can be trained with both supervised learning and reinforcement learning approaches. In this paper, we only introduce \hussein{and use} the supervised learning approach. 
\hussein{Its training loss is defined as} \yuxuan{$L_{DH} = \frac{w_{\mathit{safe}}}{N_{\mathit{safe}}}\sum_{x_{\mathit{safe}}}\sigma\big(\epsilon_{\mathit{safe}}-a_\theta(x_{\mathit{safe}})u_{\mathit{safe}}+b_\theta(x_{\mathit{safe}})\big)+ \frac{w_{\mathit{unsafe}}}{N_{\mathit{unsafe}}}\sum_{x_{\mathit{unsafe}}}\sigma\big(\epsilon_{\mathit{unsafe}}+a_\theta(x_{\mathit{unsafe}})u_{\mathit{unsafe}}-b_\theta(x_{\mathit{unsafe}})\big),$}
where $x_{\mathit{safe}}$, $x_{\mathit{unsafe}}$, $u_{\mathit{safe}}$, and $u_{\mathit{unsafe}}$ are safe and unsafe states and actions sampled from the offline dataset.

\section{PVRs as backbones for safety filters}

In this section, we \hussein{describe our method in training vision-based safety filters with PVRs as perception backbones.}
As is shown in Figure~\ref{fig:framework}, for a vehicle with $N$ cameras, \hussein{we denote  
the set of frames captured 
at time $t$,} by 
$I^{1:N}_t$. Then, \hussein{we feed these frames to}  
the \hussein{same} 
backbone model\hussein{, one at a time,} to obtain 
\hussein{their} representations\hussein{. We denote the resulting set of representations by} $h_t^{1:N}$.
\hussein{To obtain a low-dimensional vector summarizing this set of representations, 
we define an attention layer to fuse them.}

As different cameras will capture different \hussein{parts of the environment, we need to keep track of which frame corresponds to which camera when fusing the representations. To do that,}
we concatenate the representation \hussein{of each frame} with 
\hussein{a} position encoding \hussein{identifying the camera that captured it.}
We denote the resulting set of position encoding-annotated representations by ${h_t^{\prime}}^{1:N} := \hussein{\{h_t^i\ ||\ \mathrm{POS}(i)\}_{i\in [1:N]}}$,
where $||$ is the 
concatenation operation, $\mathrm{POS}$ is the position encoding function where the input is the index of the camera. In our experiments, the position encoding is just the one-hot encoding of each camera.
\hussein{The attention layer computes} the score of the representation ${h_t^{\prime}}^{i}$ of \hussein{the frame captured by} camera $i$ with $score({h_t^{\prime}}^{i}) := W^i{h_t^{\prime}}^{i}$,
where $W^i$ \hussein{is a matrix with}  
trainable parameters. 
Then, \hussein{it aggregates the results using a weighted sum with $h^*_t := \sum_{i=1}^N score({h_t^{\prime}}^{i}){h_t^{\prime}}^{i}$.} 
\hussein{Since the frames at} a single time step \hussein{do not capture critical parts of the state, such as velocity and acceleration,} 
we 
\hussein{define} the system state \hussein{$x_t$ at time $t$ to be a function of the fused representation $h^*_t$, the state at $t-1$, and the control followed at $t-1$.} 
\hussein{We consider that function to be a trainable feedforward} 
neural network. Hence, $x_t = \mathrm{MLP}_\theta(h^*_t,x_{t-1},u_{t-1})$,
where $\mathrm{MLP}_\theta$ is a \hussein{multi-level perceptron} 
parameterized with $\theta$. We set $x_{-1}$ to zero and $u_{-1}$ to $u_0$.

\section{Experimental Results}
\label{sec:result}
\begin{table*}[h]
    \centering
    \caption{Performance of the filters when trained using different  methods and on top of different backbone encoders.}
    \begin{tabular}{l|l|lll|lll}
    \toprule
    Method & Backbone & Safe States & Unsafe States & AUC (States) & Safe Actions & Unsafe Actions & AUC (Actions)\\
    \midrule
    \multirow{5}{*}{iDBF}&ViT&\textbf{96.92$\pm$0.73}&38.70$\pm$7.36&83.87$\pm$1.12&\textbf{97.24$\pm$0.67}&23.04$\pm$5.01&78.54$\pm$1.43\\
    &CLIP&93.56$\pm$3.19&55.30$\pm$7.48&83.6$\pm$2.10&94.06$\pm$3.09&38.53$\pm$8.71&78.57$\pm$1.90\\
    &VC-1&91.67$\pm$3.28&\textbf{60.61$\pm$9.48}&\textbf{87.22$\pm$1.58}&92.36$\pm$3.13&\textbf{44.51$\pm$10.18}&\textbf{82.96$\pm$1.47}\\
    &ResNet&95.66$\pm$1.36&28.00$\pm$5.32&75.35$\pm$2.25&96.18$\pm$1.3&19.73$\pm$4.61&71.25$\pm$2.56\\
    \cmidrule{2-8}&GT&90.8$\pm$1.33&67.57$\pm$2.79&85.71$\pm$1.26&91.65$\pm$1.33&49.62$\pm$3.05&76.87$\pm$1.77\\
    \midrule
    \multirow{5}{*}{SABLAS}&ViT&\textbf{97.37$\pm$0.55}&26.26$\pm$3.08&74.90$\pm$3.17&\textbf{97.40$\pm$0.61}&17.88$\pm$3.18&51.97$\pm$0.95\\
    &CLIP&91.30$\pm$3.50&57.83$\pm$8.08&82.08$\pm$1.15&90.66$\pm$4.36&45.87$\pm$10.11&65.82$\pm$5.09\\
    &VC-1&86.34$\pm$4.82&\textbf{63.04$\pm$7.50}&\textbf{84.30$\pm$0.55}&86.73$\pm$4.25&\textbf{53.64$\pm$9.04}&\textbf{69.68$\pm$4.66}\\
    &ResNet&92.95$\pm$1.64&27.91$\pm$2.05&69.78$\pm$2.6&93.38$\pm$1.15&20.16$\pm$1.2&52.11$\pm$2.00\\
    \cmidrule{2-8}&GT&89.47$\pm$1.54&41.39$\pm$5.54&73.13$\pm$3.96&80.7$\pm$2.18&53.32$\pm$1.67&67.67$\pm$2.75\\
    \midrule
    \multirow{5}{*}{DH}&ViT&NA&NA&NA&\textbf{59.84$\pm$22.86}&70.76$\pm$14.49&68.34$\pm$4.22\\
    &CLIP&NA&NA&NA&34.95$\pm$7.76&82.72$\pm$8.85&65.22$\pm$2.98\\
    &VC-1&NA&NA&NA&26.84$\pm$15.39&\textbf{91.79$\pm$5.47}&69.46$\pm$5.79\\
    &ResNet&NA&NA&NA&56.38$\pm$10.35&70.82$\pm$12.79&\textbf{69.69$\pm$4.80}\\
    \cmidrule{2-8}&GT&NA&NA&NA&91.65$\pm$1.33&49.62$\pm$3.05&76.87$\pm$1.77\\
    \bottomrule
    \end{tabular}
    \label{tab:backbone}
\end{table*}

To assess how the backbones influence the performance of the \hussein{safety filters, we conduct experiments with four different PVRs and the three training methods mentioned earlier on the DeepAccident dataset \cite{deepaccident}.}
 
 DeepAccident is a synthetic autonomous driving dataset generated using CARLA \cite{Carla}, simulating a variety \hussein{of real accidents documented by the National Highway Traffic Safety Administration (NHTSA). }
 The dataset includes \hussein{action-annotated videos}
 captured by cameras positioned at six different locations \hussein{on the ego vehicle}, as well as LiDAR data, along with ground truth position and velocity data. We only use the image data for training and evaluation. The control input to the dynamical models we train is the ego vehicle's linear velocity. 


We evaluate the performance of \hussein{the filters we train on the following metrics:} (a) distinguishing between safe and unsafe states \hussein{by assigning the CBF non-negative values for the latter and negative ones for the former, as required in (\ref{eq:cbf}), and}
\hussein{
(b) distinguishing controls that belong to the set in (\ref{eq:CBF_set}), i.e., satisfy the third condition for CBFs in (\ref{eq:cbf}), from those which do not.} 
\hussein{Note that trivial filters can achieve perfect classification accuracy of safe states (or controls) at the cost of having a zero classification accuracy for unsafe ones, and vice versa. Hence, to have a more informative measure,}
 we report the ROC-AUC scores on both of these state and control classification tasks. For each filter training  method and PVR backbone pair, we repeat the experiment with five random seeds and report the average and variance of its performance.

\subsection{Data processing}

Since the test set of DeepAccident does not include vehicle velocity data, we used its validation set as our test set, and 
\hussein{used} 10\% of its training set as our validation set. Also, it does not provide the time of collisions in the trajectories, so we manually labeled them. 
For each accident trajectory, we identified the first frame at which the collision happens and label all subsequent states \hussein{and controls} as \hussein{unsafe.}
Additionally, we \hussein{consider}
the states a the five time steps preceding the collision as safe  
but \hussein{the controls followed during that period as} 
unsafe. 
We consider all other frames and controls 
to be safe.
In \hussein{accident-free}
trajectories, we consider all frames and controls  \hussein{to be safe.}
\hussein{We divided} each trajectory 
into sub-trajectories, each consisting of five frames \hussein{corresponding to a 0.5 seconds interval}, which were categorized into three types: safe (safe frames and safe controls), unsafe (unsafe frames and unsafe controls), and transition (safe frames and unsafe controls). To alleviate data imbalance during training, we over-sampled all the unsafe and transition sub-trajectories by a factor of ten and used only five safe sub-trajectories from each trajectory during training.

\subsection{Which PVR is best suited for safety filters?}
We \hussein{evaluated}  
several 
PVRs 
to investigate whether a particular one results in a  superior performance with existing \hussein{methods for training neural safety filters for black-box models.}
\yuxuan{The models included in our study are ViT \cite{ViT} (pre-trained with ImageNet-21k dataset), CLIP \cite{CLIP} (pre-trained with image-text pairs collected from the internet), VC-1 \cite{vc1_2023} (pre-trained with the Embodied AI dataset known as CORTEXBENCH), and ResNet \cite{resnet} (pre-trained with ImageNet-1k dataset). We use the ViT-B/32 architecture for ViT, CLIP, and VC-1 and the ResNet50 architecture for ResNet. Besides PVRs, we also add one baseline (GT) that uses the ground truth information of the environment (the location and velocity of all the agents in the task) as its input.}


\begin{table*}[h]
    \centering
    \caption{Performance of the filters when the state is computed using fused and unfused representations of  camera feeds.}
    \begin{tabular}{l|l|lll|lll}
    \toprule
    Method & State & Safe States & Unsafe States & AUC (States) & Safe Actions & Unsafe Actions & AUC (Actions)\\
    \midrule
    \multirow{2}{*}{iDBF}&Fused&96.92$\pm$0.73&\textbf{38.70$\pm$7.36}&\textbf{83.87$\pm$1.12}&97.24$\pm$0.67&\textbf{23.04$\pm$5.01}&\textbf{78.54$\pm$1.43}\\
    &Unfused&\textbf{97.56$\pm$1.38}&29.13$\pm$15.53&71.22$\pm$3.85&\textbf{97.85$\pm$1.17}&19.24$\pm$10.65&65.08$\pm$3.29\\
    \midrule
    \multirow{2}{*}{SABLAS}&Fused&\textbf{97.37$\pm$0.55}&26.26$\pm$3.08&\textbf{74.90$\pm$3.17}&\textbf{97.40$\pm$0.61}&17.88$\pm$3.18&\textbf{51.97$\pm$0.95}\\
    &Unfused&50.01$\pm$49.99&\textbf{50.00$\pm$50.00}&49.65$\pm$0.70&50.00$\pm$50.00&\textbf{50.00$\pm$50.00}&50.45$\pm$0.82\\
    \midrule
    \multirow{2}{*}{DH}&Fused&NA&NA&NA&\textbf{63.58$\pm$19.47}&61.82$\pm$17.11&66.55$\pm$5.53\\
    &Unfused&NA&NA&NA&18.6$\pm$9.27&\textbf{87.72$\pm$8.27}&\textbf{66.85$\pm$3.44}\\
    \bottomrule
    \end{tabular}
    \label{tab:fuse}
\end{table*}

Table~\ref{tab:backbone} presents the performance of various backbone models across the three safety filter \hussein{training} methods. It is important to note that the state classification accuracy for the \hussein{DH method}
is not calculated, as its filters directly output hyperplanes \hussein{that classify} 
controls but not states. As shown in Table~\ref{tab:backbone}, no single model consistently outperforms others across all safety filter methods. Surprisingly, GT only achieves the best AUC with DH, while VC-1 outperforms the ground truth on state/action AUC scores with iDBF and SABLAS. This might be because VC-1 benefits from the rich semantic information present in images beyond locations and velocities of other agents, such as their poses. It is also trained with datasets from control benchmarks, equipping it with physical priors. For iDBF and SABLAS, VC-1 and CLIP perform better than ViT and ResNet. Also surprisingly, ResNet has the highest AUC with DH, although its unsafe control classification accuracy is the lowest. We also observe that models with the ViT backbone classify most states and controls as safe, performing poorly on classifying unsafe ones. 
We can see similar results on ResNet, which is trained on the same dataset. 
On the contrary, VC-1 performs better in unsafe states/actions. Particularly when using the DH method, VC-1 achieves the highest accuracy (91.79\%) on classifying unsafe actions. When using other methods, VC-1 also achieves the best unsafe state/action classification accuracy and AUC scores on action classification. This is probably because VC-1 is trained on multiple robotic datasets, including navigation datasets where there are lots of unsafe/collision scenes. Models with VC-1 backbone are more capable of detecting unsafe states. 
Similarly, CLIP is trained on data collected from the internet. There might be data related to traffic or collision scenes in its training dataset that enhance its capability of processing safety related features. 
Based on the above analysis, we conclude that backbones pre-trained with data related from control or robotics domains  lead to a better performance.

\subsection{Which safety filter method is better for \hussein{vision-based}
autonomous driving?
}
We can observe that both CBF-based methods perform poorly in detecting unsafe actions. The highest unsafe action classification accuracy of CBF-based methods (53.64\%), which is lower than that the lowest of DH (70.76\%). This might be due to the errors in the learned dynamical models in the former methods. In the DeepAccident dataset, there are significantly fewer accident sub-trajectories compared to regular sub-trajectories, 
 which means that the trained dynamical models may perform poorly when it comes to unsafe scenarios, affecting the final performance of the filters. However, DH \hussein{may be less vulnerable to}
this issue as it \hussein{circumvents the need for a dynamical model and directly learns state-dependent control classifiers.}
DH achieves the best accuracy on unsafe actions classification (91.79\%). However, DH 
performs poorly on safe action detection. Even the highest accuracy of DH doing that is lower than the lowest accuracy of iDBF and SABLAS. 

\subsection{\hussein{Shortcomings of the dataset and its labels}}
In the experiment, we found that barriers trained with the 
\hussein{iDBF} and SABLAS methods were \hussein{less effective}
in serving as \hussein{safety filters.}
When used as constraints in quadratic programming, the control input contributed minimally due to the inaccurate dynamics learned through these methods. For the constraint in (\ref{eq:qp}), $\nabla B(x)f(x) + \gamma(B(x))$ was much larger than $\nabla B(x)g(x)u$ and dominated the whole term, which means that the satisfaction of the constraint depends on the state instead of the control input the controller chooses.
Interestingly, the \hussein{DH method}
\hussein{led to safety filters that are able to detect}
unsafe \hussein{controls at safe and unsafe states,}
although the classification accuracy of safe controls was not perfect.
However, it \hussein{is not as effective in detecting} 
safe controls. These limitations are likely due to the shortcomings of the 
dataset and the data annotation. Since there is only one control followed at every state in a trajectory, the model 
cannot learn to separate between unseen safe and unsafe controls under the same state. Moreover, we labelled all the controls in the non-accident trajectories as safe. However, this might not be accurate, as unsafe controls that did not lead the vehicle to collision would have been labelled safe. Also, our labelling of the five frames before the collision as safe frames under unsafe controls might not be accurate as these states might already be unsafe if very close to collision and it might not be that older control inputs have been also unsafe. 

\subsection{Is it better to have states based on fused or unfused representations of the frames from the multiple cameras?}
We experiment with both approaches and show our results in \hussein{Table}~\ref{tab:fuse}. 
For the fused case, we train an attention layer after the backbone model to generate a system state that fuse the representations of the frames from all the cameras. In this case, the dynamical model is defined over the 
the state that is a function of the fused representations. 
For the unfused case, a separate state is defined based on each camera. The composition of all the states represent the system state. 
We train the dynamical model to generate the trajectories of  the states corresponding to each camera separately. 
We then train an attention layer that fuse the states before given as input to the safety filter. In this experiment, we use ViT as the backbone model for each filter training method.

As shown in 
\hussein{Table}~\ref{tab:fuse}, we can observe that models 
achieve higher AUC over both states and controls with the CBF-based filter training methods when fusing representations to define the state. With DH, although the AUC and unsafe controls classification accuracy of the unfused case is slightly higher than that of the fused one, its safe controls classification accuracy is low. 
Thus, we can still say that the fused representation performs better than the unfused one with that training method too. 
With iDBF and SABLAS, in the unfused case, the dynamical model has to be able to generate trajectories of states based on the representations of frames from different cameras, which can be a more complex task compared to the fused case, where it only needs to generate the trajectories of the state based on the fused representation. For DH, the improvement may be brought by the more concise system state. 
As the system state is formed by the representation of current observation along with the system state and action of the last time step, an unfused history system state may contain redundant information and mislead the current system state to hold redundant information.

 
\section{Conclusion} 
\yuxuan{In this paper, we trained vision-based safety filters for collision avoidance in autonomous driving . We used four different frozen PVRs as perception backbones and experimented with two approaches for defining the state of the vehicle and its environment. 
We used three different existing methods for training safety filters for black-box dynamics. We compared the performance of the filters resulting from the different combinations of backbones and training methods, showing the competitiveness of our approach with respect to non-vision-based filters with access to the ground truth state. Our future plans include evaluating our models online  to confirm our offline  results. They also include addressing the uncertain dynamics problem that result from training  models that are independent from the actions of surrounding vehicles. 
} 
\label{sec:conclusion}


\clearpage


\bibliographystyle{IEEEtran}
\bibliography{references,Hussein,Yuxuan}

\end{document}